\documentclass{article} % For LaTeX2e
\usepackage{iclr2027_conference,times}

\usepackage{amsmath,amsfonts,bm}

\def\eqref#1{equation~\ref{#1}}
\def\1{\bm{1}}

\DeclareMathAlphabet{\mathsfit}{\encodingdefault}{\sfdefault}{m}{sl}
\SetMathAlphabet{\mathsfit}{bold}{\encodingdefault}{\sfdefault}{bx}{n}

\usepackage{hyperref}
\usepackage{url}
\usepackage{graphicx}
\usepackage{subcaption}
\usepackage{booktabs}
\usepackage{siunitx}
\usepackage{amssymb}
\usepackage{amsfonts}
\usepackage{amsmath}
\usepackage{makecell}
\usepackage{amsthm}
\usepackage{tabularx}
\usepackage{colortbl}
\usepackage{capt-of}
\usepackage{ragged2e}

\usepackage{algorithm}
\usepackage{algpseudocode}
\usepackage{tabularx}
\theoremstyle{plain}

\theoremstyle{definition}

\theoremstyle{remark}

\usepackage{adjustbox}

\usepackage{wrapfig}
\title{ASPEN: Adaptive Semantic Reasoning via Command Persistence Prediction for Efficient Navigation VLAs}

\author{Vieri N.~Akustia\\
Department of Electrical and Computer Engineering\\
University of Houston \\
Houston, TX 77004, USA \\
\texttt{vnakustia@cougarnet.uh.edu} \\
\And
Bin Hu \\
Departement of Engineering Technology \\
University of Houston \\
Houston, TX 77004, USA \\
\texttt{bhu11@central.uh.edu} \\
}
\iclrfinalcopy % Uncomment for camera-ready version, but NOT for submission.
\begin{document}
\raggedbottom

\maketitle

\begin{abstract}
Navigation vision-language-action (VLA) policies repeatedly invoke costly semantic reasoning even when the next command remains unchanged. Command reuse can reduce computation but risks missing instruction-dependent turns or stops. We introduce ASPEN, an event-triggered scheduler based on \textbf{command persistence prediction}: estimating whether the frozen base policy would reissue its previous command. A lightweight multimodal predictor combines temporal observations, language instructions, execution history, and cached VLA confidence. At eligible command boundaries, sufficient predicted persistence permits command reuse. Otherwise, ASPEN triggers fresh semantic reasoning from the VLA. Evaluated with frozen NaVILA on cross-dataset RxR-CE val-unseen, ASPEN reduces VLA calls by \textbf{9.5\%} and estimated inference time by \textbf{8.1\%}, including scheduling overhead. It maintains comparable navigation performance and preserves identical baseline action sequences on \textbf{92.1\%} of episodes.
\end{abstract}

%===============================================================================

\section{Introduction}
\begin{figure}[H]
    \centering
    \includegraphics[width=\linewidth]{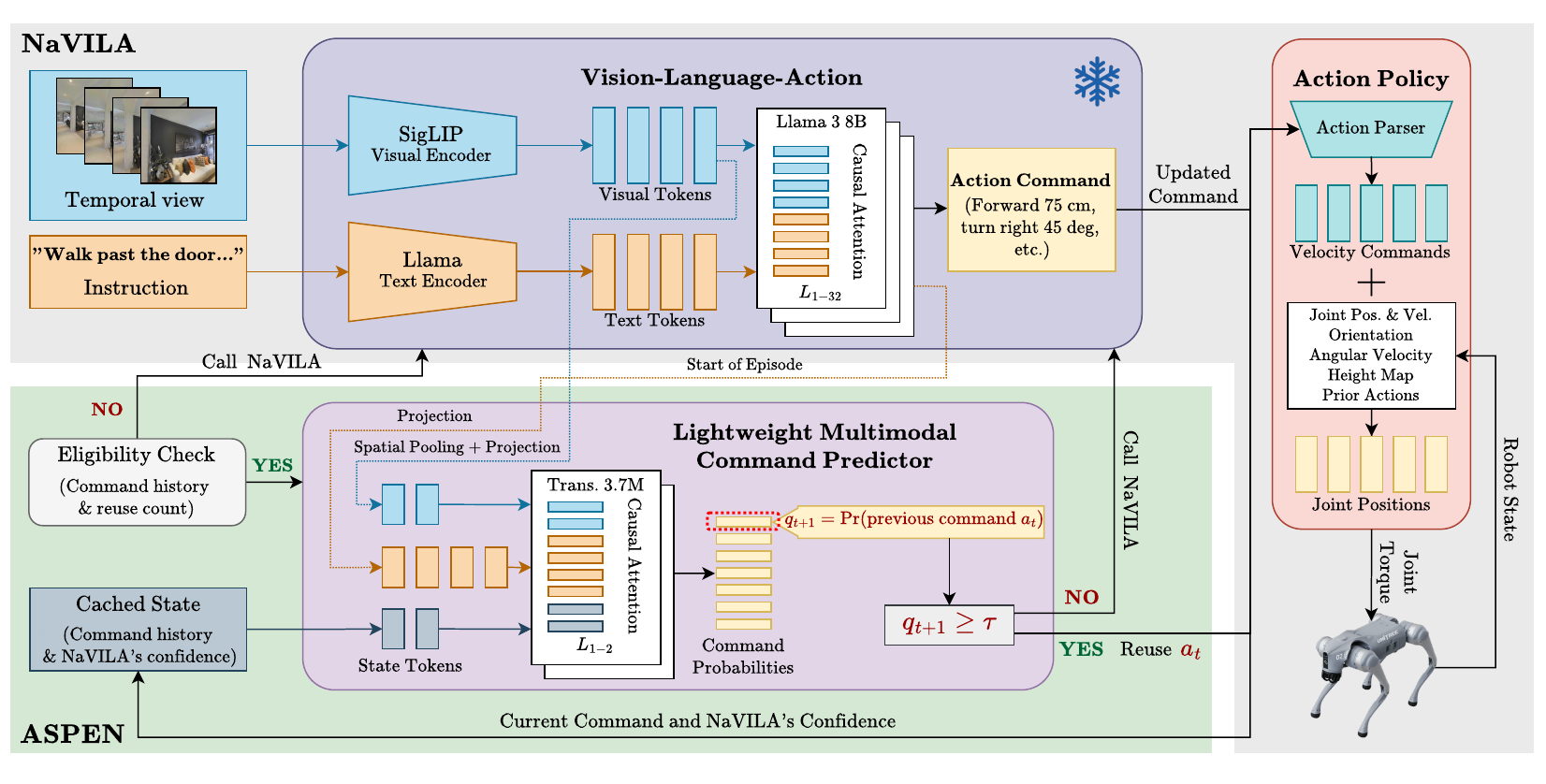}
    \caption{ASPEN schedules a frozen navigation VLA. A learned persistence score selects command reuse or fresh VLA inference, preceded by an eligibility check.}
    \label{fig:overview}
\end{figure}

Vision-language-action (VLA) policies connect visual observations and language instructions to robot behavior~\cite{palme, rt2, openvla}. In navigation, these policies can generate high-level commands that are executed through a downstream action interface~\cite{waypoint, etpnav, navillm, navid, navila}. However, repeated invocation of a large VLA incurs substantial cost when successive decisions produce the same command. We study when this semantic reasoning can be omitted while retaining the behavior of a frozen navigation policy.

Existing efficiency methods use action-sequence prediction~\cite{act, diffusionpolicy, octo}, selective computation and compression~\cite{molevla, efficientvla}, temporal caching~\cite{vlacache, gatedcache}, and adaptive execution~\cite{flashvla, spvla, ffdcwam}. FlashVLA~\cite{flashvla} combines action-history consistency and visual-token overlap to trigger reuse, whereas SP-VLA~\cite{spvla} schedules between a VLA and a lightweight action generator using motion history. In navigation, substantial visual change can accompany continued forward motion, while the current view alone may not establish whether an instructed landmark has already been passed. The relevant scheduling question is therefore not simply whether the scene or motion has changed, but whether the base policy's next command would change given the instruction and execution history.

We introduce ASPEN, which predicts this \emph{command persistence} from temporal observations, language, execution history, and cached VLA confidence. Following the event-triggered principle~\cite{tabuada2007, heemels2012}, the scheduler chooses between repeating the previous command and refreshing NaVILA. Unlike a lightweight replacement controller, its predicted command distribution is used to schedule the base policy, not to select a different executable command. Our contributions are \textbf{(i)} command-persistence-based scheduling for a frozen navigation policy, \textbf{(ii)} a context-conditioned predictor with explicit eligibility check, and \textbf{(iii)} cross-dataset evaluation from R2R-CE to RxR-CE~\cite{vlnce, rxr}, demonstrating reduced estimated inference cost with similar observed success and high action-sequence agreement.

%===============================================================================

\section{ASPEN: Command Persistence Prediction}
\label{sec:method}
ASPEN operates at high-level command boundaries, not individual simulator steps (Figure~\ref{fig:overview}). At boundary $t+1$, the preceding command $a_t$ has completed and observation $o_{t+1}$ is available. Before feature extraction, an eligibility check $e_{t+1}$ restricts predictor invocation to parsed forward commands with fewer than eight consecutive reuses.

\textbf{Lightweight multimodal command predictor.} Frozen SigLIP encodes the episode's first view and the two most recent historical decision-point views. Each view is pooled into a $9\times9$ token grid and projected to width 256. Frozen Llama-3 encodes the instruction text once per episode, projecting its final hidden states to the same width. For each historical decision, a learned state vector encodes the executed command, magnitude, step count, repeat flag, episode-progress features, and cached NaVILA confidence. Confidence is refreshed only when NaVILA is invoked. Following DINO-WM~\cite{dinowm}, the state vector is added to the corresponding visual tokens and also included as a separate state token.

Two width-256 transformer layers with four attention heads apply causal self-attention over visual and state tokens, instruction cross-attention~\cite{lcvn}, and feed-forward processing. Instruction tokens serve as keys and values. The resulting state output $h_t$ queries the separately encoded current view $o_{t+1}$ through attention pooling. The current view's 81 tokens do not pass through the transformer. An MLP outputs $p_\theta(\cdot\mid o_{t+1},h_t)$ over NaVILA's ten canonical commands.

\paragraph{Command-persistence trigger.}
Let $\mathcal A_{\mathrm{fwd}}$ denote eligible forward commands and $r_t$ the consecutive reuse count before boundary $t+1$. The scheduler uses
\begin{equation}
\begin{aligned}
e_{t+1}&=\mathbf 1\{a_t\in\mathcal A_{\mathrm{fwd}}\ \land\ r_t<8\},
& q_{t+1}&=p_\theta(a_t\mid o_{t+1},h_t),\\[-1pt]
a_{t+1}&=
\begin{cases}
a_t, & e_{t+1}=1\ \land\ q_{t+1}\geq\tau,\\
a^{\mathrm{NaVILA}}_{t+1}, & \text{otherwise}.
\end{cases}
\end{aligned}
\label{eq:persistence_trigger}
\end{equation}
The persistence score $q_{t+1}$ is the learned probability mass on the previous canonical command and is computed only when $e_{t+1}=1$. Ineligible boundaries invoke NaVILA directly. The reference command $a^{\mathrm{NaVILA}}_{t+1}$ is generated from the current execution history only on a refresh, not to verify reuse. Both branches share the downstream execution interface, and NaVILA's observation history continues updating during reuse. This score predicts reference-command agreement rather than certifying task correctness.

\textbf{Training and trigger calibration.} We train ASPEN with ten-class next-command cross-entropy on frozen NaVILA rollouts. The R2R-CE training split is divided into disjoint training, checkpoint-selection, and calibration sets: 42/6/13 buildings and 2,536/405/1,685 episodes, respectively. Training includes additional examples collected after incorrect command reuse. We use AdamW with learning rate $5\times10^{-4}$ and batch size 32, training three seeds for up to 100 epochs with 15-epoch early-stopping patience. For each seed, we retain the checkpoint with the lowest checkpoint-selection loss. Deterministic replay on the calibration set selects the threshold maximizing call savings subject to an empirical divergence rate of at most $10\%$. We deploy the median seed by replay savings, yielding $\tau=0.9418$, and evaluate it without RxR-specific tuning. This source-domain selection criterion provides no target-domain divergence guarantee.

%===============================================================================

\section{Experimental Results}
\label{sec:experiments}
\begin{table}[t]
\centering
\caption{Cross-dataset evaluation on 1,223 RxR-CE val-unseen episodes. Parentheses show changes relative to baseline NaVILA.}
\label{tab:rxr}
\fontsize{9}{11}\selectfont
\setlength{\tabcolsep}{3pt}
\renewcommand{\arraystretch}{1.0}

\begin{tabularx}{\linewidth}{@{}>{\raggedright\arraybackslash}Xccccccc@{}}
\toprule
\makecell[c]{Method} &
\makecell[c]{VLA calls/ep\\(vs baseline)} &
\makecell[c]{Inference s/ep\\(vs baseline)} &
\makecell[c]{SR $\uparrow$} &
\makecell[c]{OS $\uparrow$} &
\makecell[c]{nDTW $\uparrow$} &
\makecell[c]{NE $\downarrow$} &
\makecell[c]{Exact\\agreement (\%) $\uparrow$} \\
\midrule
Baseline NaVILA &
75.9 &
41.5 &
53.0 &
59.9 & 63.3 & 5.93 &
100.0\%\\
\midrule

FlashTrigger \cite{flashvla} &
72.0 ($-5.1\%$) &
39.7 ($-4.5\%$) &
53.0 &
\textbf{59.9} & \textbf{63.3} & \textbf{5.93} &
\textbf{\boldmath 97.5\%}\\

SP-VLA \cite{spvla} &
69.4 ($-8.5\%$) &
38.0 ($-8.5\%$) &
42.7 &
53.2 & 53.2 & 7.17 &
7.3\%\\

\rowcolor{yellow!15}
Free Rule (ours) &
\textbf{\boldmath 61.0 ($-19.6\%$)} &
\textbf{\boldmath 33.4 ($-19.6\%$)} &
49.1 &
57.3 & 58.4 & 6.64 &
7.1\%\\

\rowcolor{yellow!15}
Full Gate (ours) &
66.9 ($-11.7\%$) &
40.2 ($-3.3\%$) &
51.6 &
59.5 & 61.8 & 6.36 &
21.8\%\\

\rowcolor{orange!15}
ASPEN (ours) &
68.7 ($-9.5\%$) &
38.2 ($-8.1\%$) &
\textbf{53.1} &
59.8 & 63.2 & 5.95 &
92.1\%\\
\bottomrule
\end{tabularx}
\end{table}

\textbf{Evaluation and comparators.} We evaluate the selected scheduler on 1,223 RxR-CE val-unseen episodes without target-dataset training or tuning. Comparators include frozen NaVILA, navigation adaptations of FlashVLA's FlashTrigger~\cite{flashvla} and SP-VLA~\cite{spvla} using published default thresholds, and two diagnostic alternatives. Free Rule repeats a preceding 75\,cm forward command once. Full Gate uses a recurrent classifier, with threshold $0.70$, after each forward command to choose between an additional 25\,cm forward step and a NaVILA call.

We report success rate (SR), oracle success (OS), normalized dynamic time warping (nDTW), and navigation error (NE) as navigation performance metrics. Efficiency is reported by VLA calls and estimated inference time per episode. Exact agreement is the percentage of episodes with the same complete executed action sequence as baseline NaVILA. This measures fidelity to the reference policy, not whether its trajectory succeeds.

\textbf{Inference-cost estimation.} We combine logged invocation counts with fixed unit costs measured on a dedicated NVIDIA RTX PRO 6000 Blackwell GPU over 241 ASPEN gate calls across 20 profiling episodes. ASPEN's estimated gate cost is 19.45\,ms, the sum of component-wise median costs for image decoding, preprocessing, visual encoding, pooling, prediction, and communication. The predictor alone accounts for 1.5\,ms, while the shared SigLIP visual encoding accounts for 10.5\,ms. Episode-averaged estimates include scheduling overhead and once-per-episode instruction encoding, with the same estimation procedure used for all methods. Reported results are estimated inference costs, not measured end-to-end navigation runtimes.

\begin{figure}[t]
    \centering
    \includegraphics[width=\linewidth]{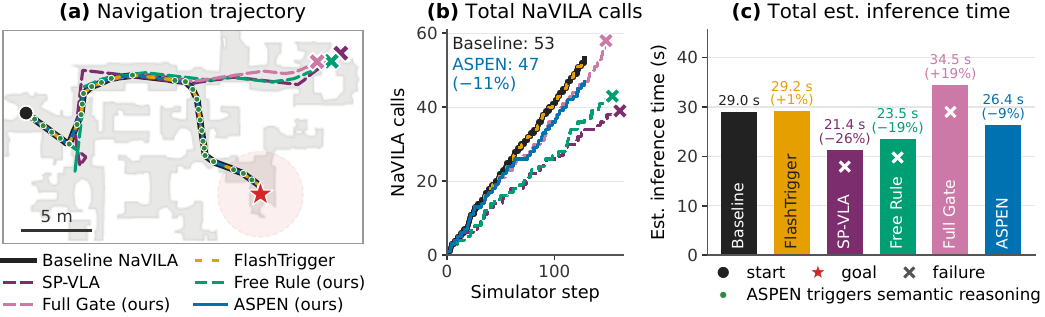}
    \caption{Example RxR-CE val-unseen episode with baseline action sequence preservation. ASPEN reduces NaVILA calls from 53 to 47 (11\%) and estimated inference time from 29.0 to 26.4\,s (9\%).}
    \label{fig:baseline_compare}
\end{figure}

\begin{figure}[t]
    \centering
    \includegraphics[width=\linewidth]{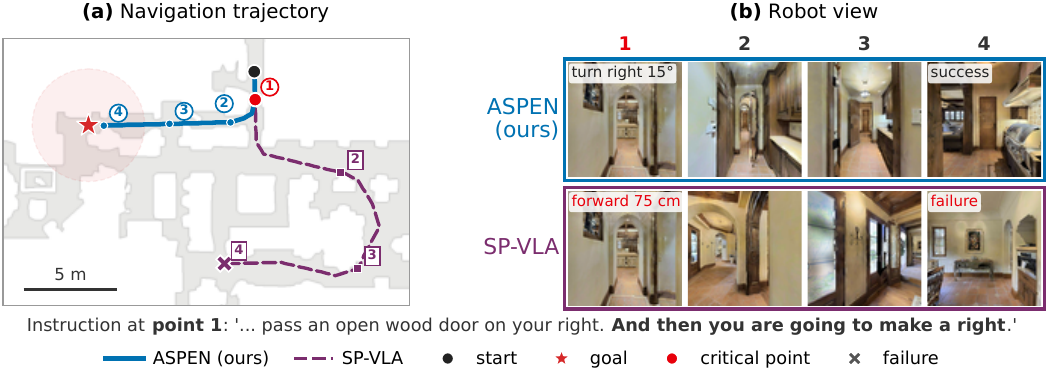}
    \caption{A critical command boundary in an example RxR-CE val-unseen episode. ASPEN refreshes semantic reasoning and turns right, whereas the navigation-adapted SP-VLA continues forward and subsequently fails.}
    \label{fig:example_trigger}
\end{figure}

\textbf{Efficiency--performance trade-off.} Table~\ref{tab:rxr} shows that ASPEN maintains comparable SR, OS, nDTW, and NE to baseline NaVILA while reducing estimated inference time by 8.1\%. Compared with FlashTrigger, ASPEN uses 3.3 fewer calls per episode, but agreement decreases from $97.5\%$ to $92.1\%$. The additional savings therefore involve an efficiency--agreement trade-off rather than improvement on navigation performance.

SP-VLA has slightly lower estimated inference time than ASPEN (38.0 versus 38.2\,s), but its success rate falls to $42.7\%$. Free Rule offers the largest call and estimated-time reductions of $19.6\%$ by utilizing its simple refresh condition. However, its success rate falls to $49.1\%$, and exact agreement to $7.1\%$. Full Gate reduces calls by $11.7\%$ yet estimated time by only $3.3\%$, emphasizing the need to include scheduling overhead. Efficiency--performance trade-off for all methods shown in Appendix \ref{app:add_details}

Figure~\ref{fig:baseline_compare} illustrates ASPEN's adaptive scheduling. It reduces unnecessary semantic reasoning while maintaining the reference trajectory, achieving successful navigation with 9\% estimated inference-time savings. Figure~\ref{fig:example_trigger} illustrates how ASPEN's context-conditioned predictor triggers semantic reasoning when necessary. At point 1, the instruction calls for a right turn, and a walkway is visible. ASPEN refreshes semantic reasoning and turns right, whereas SP-VLA continues forward.

%===============================================================================

\section{Conclusion}
\label{sec:conclusion}
ASPEN enables efficient, event-triggered semantic reasoning in navigation VLAs through command persistence prediction. Its lightweight multimodal command predictor combines navigation context with cached VLA confidence to determine when the previous command can be reused. Cross-dataset evaluation with frozen NaVILA shows reduced reasoning cost while maintaining comparable navigation performance and high agreement with baseline action sequences. Future work will examine transfer to other navigation VLAs and measure latency and energy savings on physical robots.

%===============================================================================

\clearpage

% \subsection*{Acknowledgements}
% This work was supported by the NVIDIA Academic Grant Program.

% \subsection*{AI use statement}

% In this work, we used generative AI tools to aid and polish manuscript writing and to assist with research execution through debugging and modifying portions of the research code.
% We have not used generative AI tools to generate experimental data or make final scientific conclusions, and other disclosure categories not relevant to these uses are not applicable to this work.
% We reviewed all AI-assisted work.
% AI-assisted code was manually inspected and tested for correctness, while AI-assisted text was reviewed and revised by the authors for technical accuracy and consistency with the implemented methods and reported results.
% We take responsibility for the final content of this work, including text, claims, code, and artifacts produced with the aid of generative AI.

\bibliography{iclr2027_conference}
\bibliographystyle{iclr2027_conference}

\clearpage
\appendix

\section{Additional Results}
\label{app:add_details}

\begin{figure}[H]
    \centering
    \includegraphics[width=\linewidth]{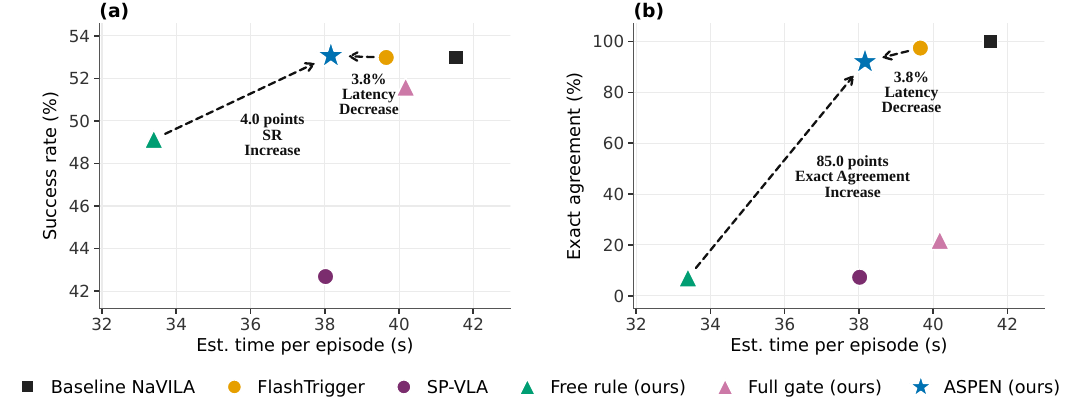}
    \par
    \caption{Efficiency--performance trade-offs on RxR-CE val-unseen. Each point represents a method in Table \ref{tab:rxr}. Both panel compare estimated inference time per episode with (a) success rate and (b) identical trajectory rate.}
    \label{fig:trade-off}
\end{figure}

Figure \ref{fig:trade-off} shows that ASPEN offers the best efficiency--performance trade-off. It offers 3.8\% more inference time savings compared to FlashTrigger while achieving 4 points higher success rate and 85 points higher exact agreement compared to Free Rule.

\end{document}